# Statistical Method to Model the Quality Inconsistencies of the Welding Process

Mohammad Aminisharifabad, Qingyu Yang

Department of Industrial and System Engineering

Wayne State University

**Abstract**: Resistance Spot Welding (RSW) is an important manufacturing process that attracts increasing attention in automotive industry. However, due to the complexity of the manufacturing process, the corresponding product quality shows significant inconsistencies even under the same process setup. This paper develops a statistical method to capture the inconsistence of welding quality measurements (e.g., nugget width) based on process parameters to efficiently monitor product quality. The proposed method provides engineering efficiency and cost saving benefit through reduction of physical testing required for weldability and verification. The developed method is applied to the real-world welding process.

1. **Introduction:**

   Advance high strength steel (AHSS) is gaining increasing attention in industries, especially in auto industry. The main reason of popularity of the steel is its unique mechanical properties. The most widely used AHSS in auto industry is two-phase (DP) AHSS.

   Resistance spot welding (RSW) process is the widely used steel joining method in auto industry. Quality of RSW strongly affects the reliability of final product. However, weldability qualification requires costly and time-consuming, which may delay selection of new materials, processes, and related designs. Figure 1 illustrates the welding process of DP AHSS in auto industry.

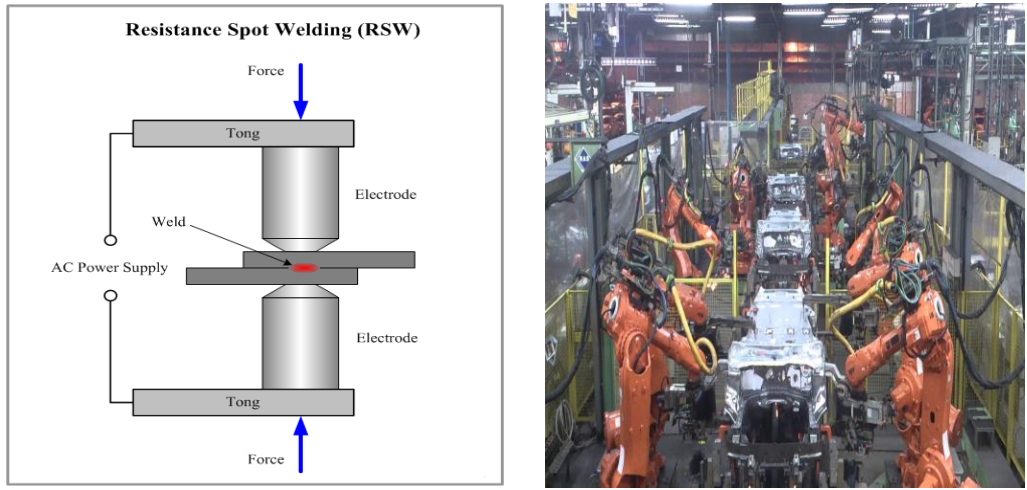

Figure 1: welding process of DP AHSS

To overcome the costly and time-consuming physical test, a machine learning-based approach is required to learning quality of welding through a training set. However, there are some challenges 1) there is some outliers in the data set 2) there is measurement error in respond variable (welding quality) 3) the dimension of data is high, i.e. many welding parameters are involved in welding process. To overcome the challenges, we proposed a novel machine learning-based model to predict welding quality of DP AHSS.

## 2. Literature review

With the large data set available, the most effective method to predict in field of data analytics is machine learning. The analytical method helps managers for decision-making. Machine learning is used successfully in many areas including bioinformatic [1-4], mechanical engineering [5] and healthcare [2]. There are many machine learning methods proposed in literature including support vector machine (SVM) [6], lasso [7], however, the exiting machine learning models cannot overcome the challenge of high dimensionally and outlier detection simultaneously.

## 3. Statistical model

The proposed model to overcome the high dimensionally and outlier detection simultaneously is formulated as follows:

$$\min \|\mathbf{w}\|$$
$$st$$
$$\sum_i [|y_i - \mathbf{w}^T x_i| \leq \delta] \geq C \times m \quad (1)$$

where $0 \leq C \leq 1$, $\delta$ are user defined constant, and $m$ is the number of training data points. $\mathbf{w} = \{w_1, ..., w_n\}$ are model parameters. In our proposed model $C$ control removing the outlier of data set and the minimizing of $\mathbf{w}$ encourage the model to be sparse i.e. only relevant parameters are selected by the model. Furthermore, noisy respond data point $y_i$ are considered to be in $y_i \pm \delta$ interval.

## 4. Parameter estimation

In order to estimate the model parameter, the model in (1) suggests that at least $C \times m$ of data points satisfy $|y - \mathbf{w}^T \mathbf{x}| \leq \delta$. Optimization problem in (1) can be solved with introducing big-M and $\lambda$ variable to transfer in to a mixed integer programming as follows:

$$\min \sum_{i=1}^{n} ||w_i||$$
$$s.t$$
$$|y_1 - \mathbf{w}^T \mathbf{x}_1| \leq \delta + M\lambda_1$$
$$\ldots$$
$$|y_m - \mathbf{w}^T \mathbf{x}_m| \leq \delta + M\lambda_m \quad (2)$$
$$\sum_{j=1}^{m} \lambda_j \leq m.C$$
$$\lambda_j = \{0,1\}$$

Since (2) is mixed integer programming, it can be solve by standard methods existing in literature [8].

## 5. Case Study

We apply our proposed model on 50 samples of AHSS to predict their welding quality. Specifically, the model predicts nugget size of the steel based on historical welding parameters.

However, we cannot apply traditional machine learning methods to predict nugget sizes since the data set has measurement error and outlier data point. Triyono, Triyono, et al [9] shows nugget size measurement is an operator depend process and has an error of 2.4%. The data has outlier due to dynamic process of spot welding.

In the case study we consider 12 welding parameters including materials (stack-up), thickness, weld force, weld current, coating and weld time. We set $\delta = 0.144$, $C = 10\%$ in the propose model.

In order to compare performance of our model with applying (SVM) and lasso on the dataset. Table 1 Shows that our proposed model outperforms the existing models.

Table 1 Performance of methods

| Method | MSE (mean square error) |
|---|---|
| SVM (non-linear) | 3.95 |
| Lasso (linear) | 4.35 |
| Sparse Robust Regression | 3.16 |

**6. Conclusion**

Welding quality highly strongly affect products, and welding qualification is a costly process. We proposed a novel robust linear model to predict quality of welding. The model can tackle measurement error, inconsistency and gross error in one stage. Furthermore, a method is developed to estimate model parameters. The methodology is validated by a case study.